\documentclass[conference]{IEEEtran}
\usepackage{times}
\usepackage{epsfig}
\usepackage{graphicx}
\usepackage{amsmath}
\usepackage{amssymb}
\usepackage[para,online,flushleft]{threeparttable}
\usepackage{makecell}
\usepackage[pagebackref=true,breaklinks=true,colorlinks,bookmarks=false]{hyperref}
\usepackage{booktabs}
\usepackage{multirow}
\usepackage[linesnumbered,ruled,vlined]{algorithm2e}
\usepackage{caption}
\usepackage[hang,flushmargin]{footmisc}
\usepackage{lipsum}

\makeatletter
\newcommand{\algorithmfootnote}[2][\small]{%
  \let\old@algocf@finish\@algocf@finish
  \def\@algocf@finish{\old@algocf@finish
    \leavevmode\rlap{\begin{minipage}{\linewidth}
    #1#2
    \end{minipage}}%
  }%
}
\makeatother

\SetKwInput{KwInput}{Input}                
\SetKwInput{KwOutput}{Output}

\begin{document}

\title{Achieving Real-Time LiDAR 3D Object Detection on a Mobile Device}

 \author{\IEEEauthorblockN{ $^1$Pu Zhao, $^2$Wei Niu, $^1$Geng Yuan, $^1$Yuxuan Cai, $^3$Hsin-Hsuan Sung, \\  $^4$Sijia Liu,  $^3$Xipeng Shen, $^2$Bin Ren, $^1$Yanzhi Wang, $^1$Xue Lin}
\IEEEauthorblockA{
\textit{$^1$Northeastern University,  Boston, MA}\\
\textit{$^2$William \& Mary, Williamsburg, VA}\\
\textit{$^3$North Carolina State University, Raleigh, NC}\\
\textit{$^4$  Michigan State University, East Lansing, MI}\\
}
}

\maketitle

\begin{abstract}
3D object detection is an important task, especially in the autonomous driving application domain.
However, it is challenging to support the real-time performance with the limited computation and memory resources on edge-computing devices in self-driving cars.
To achieve this, we propose a  compiler-aware unified framework incorporating network enhancement and pruning search with the reinforcement learning techniques, to enable real-time inference of   3D object detection on the resource-limited edge-computing devices. 
Specifically, a generator Recurrent Neural Network (RNN) is employed to provide the unified scheme for both network enhancement and pruning search automatically, without human expertise and assistance. And the evaluated performance of the unified schemes can be fed back to train the generator RNN. 
The experimental results demonstrate that the proposed framework firstly achieves real-time 3D object detection on mobile devices (Samsung Galaxy S20 phone) with competitive detection performance. 
\end{abstract}

\section{Introduction}
There are growing interests in  two aspects of the machine learning field. 
First, various Deep Neural Network (DNN) architectures have served as the fundamental building blocks of machine learning applications due to the superior  accuracy performance \cite{goodfellow2016deep}.
Second,  the edge-computing devices such as embedded systems, FPGAs, ASIC chips have been serving as the primary carriers of machine learning applications including autonomous driving, wearable health devices, video streaming, etc.~\cite{philipp2011sensor,lane2015early}. 
It is desirable and challenging to deploy DNN models on edge devices, targeting for real-time inference performance. 
Take the 3D object detection on self-driving cars as an example.
To ensure driving safety, the 3D detection with point clouds \cite{martin2019autonomous} should satisfy the real-time inference requirement when executed on resource-limited GPU devices equipped on self-driving cars, since the more powerful high-end GPUs are too costly and power-hungry to use on vehicles.
However, many state-of-the-art DNN models such as VGG-16~\cite{simonyan2014very} and Yolo \cite{bochkovskiy2020yolov4} are computation and memory-intensive for the light-weighted edge computing devices to obtain real-time performance.


To reduce the computation and memory requirements and therefore accelerate the DNN inference execution, DNN model pruning techniques have been used to remove the redundancy in DNN models \cite{wen2016learning,guo2016dynamic,he2018amc,he2019filter}. 
Various DNN pruning frameworks explore i) different pruning methods (how to prune) such as magnitude pruning \cite{frankle2018lottery} and ADMM pruning \cite{zhang2018systematic}, ii) pruning types (what to prune) such as pattern pruning \cite{ma2019pconv} and filter pruning \cite{he2019filter}, and iii) pruning ratio (how much to prune) \cite{guo2016dynamic}.

A noteworthy direction is the pattern pruning \cite{ma2019pconv}, which prunes kernels according to a predefined pattern set to facilitate compiler optimizations towards higher parallelism for inference execution and therefore superior acceleration.
There have been other  DNN inference acceleration frameworks on edge devices with compiler optimizations 
 such as Tensorflow-Lite \cite{TensorFlow-Lite}, TVM \cite{chen2018tvm}, MNN \cite{Ali-MNN}, and PyTorch \cite{paszke2019pytorch}, although they focus on unpruned models.
Moreover,  Winograd \cite{barabasz2019winograd,lavin2016fast} could  improve the inference speed by transforming tiles of the input and kernel into a modulo polynomials Winograd domain, although the transformations  cause certain overhead which may degrade the improvement. 
This paper makes an observation from all these above-mentioned frameworks that kernel size plays an important role and should be identified as a new network optimization dimension, and therefore proposes to perform the kernel size fine-tuning and adopt Winograd as the key network enhancement techniques to facilitate compiler optimization for inference acceleration.

There are several problems with the  existing efforts. First, previous pruning frameworks usually fix the pruning method and pruning type for all the layers in the model 
without customizing the best-suited pruning type for each layer.  Second, there lacks  a method to evaluate the inference speed performance of different network enhancements  in presence of
compiler optimizations.
Third,  the process of model pruning, network enhancements, and compiler optimization co-design relies heavily on professional expertise and experience, and involves large manual hyperparameter fine-tuning efforts.

On the other hand, Automated Machine Learning (AutoML) \cite{he1908automl,sebastiani2002machine} has gained ever-increasing interests and attention that tries to automatically solve  problems without human assistance under the given computational budget. In AutoML, Neural Architecture Search (NAS) \cite{zoph2016neural,liu2018darts,kandasamy2018neural,ru2020neural} receives notable attention that  designs novel network architectures without human expertise that can rival the best human-invented architectures in terms of testing accuracy.
NAS makes the architecture design less dependent on human experts   with  a solid understanding of deep learning as well as the application domain.

To satisfy the real-time inference requirement for 3D objection detection with point clouds on edge devices, and inspired by  NAS,  we propose a compiler-aware unified framework incorporating network enhancement and pruning search with reinforcement learning (RL). The framework automatically generates various unified schemes with a list of network enhancement and pruning actions. Then the performance of the models derived under the unified schemes can be fed back to the generator to maximize the expected rewards. Although there are some other DNN inference acceleration frameworks with compiler optimizations such as  Tensorflow-Lite \cite{TensorFlow-Lite}, TVM \cite{chen2018tvm} and  MNN \cite{Ali-MNN}, we are the first to provide support for real-time 3D object detection  on mobile GPUs (we use Samsung Galaxy S20 phone here).
The contributions are summarized as follows.
\begin{itemize}
  \item \textbf{Unified framework.} We propose a unified framework incorporating network enhancement and pruning search with reinforcement learning for 3D object detection. For the first time, we  provide support for real-time 3D object detection  on mobile GPUs with compiler optimization.
\item \textbf{Flexible configuration.} Distinctive from existing works using fixed pruning or network enhancement strategy, our framework enjoys the great flexibility that can be customized and optimized down to the layer level while covering state-of-the-art practices.
  \item \textbf{Compiler awareness.} The framework is able to take into account the effects of compiler optimizations during the search space exploration, through the automatic code generation of compiler with the supports of various pruning and network enhancement techniques.
   \item \textbf{Real-time performance.} To satisfy the inference speed requirement, we take the real-time performance constraint into considerations during the search space exploration as a reward for different schemes. Our experimental results demonstrate that  the proposed framework can achieve real-time 3D object detection on  mobile devices (Samsung Galaxy S20 phone). 
\end{itemize}

\section{Related Work}

\subsection{Network Enhancement}

Compiler optimization is able to  improve the inference speed by utilizing hardware parallelism more efficiently. Thus a new optimization dimension, i.e., kernel size, is introduced as different kernel size of convolutional (CONV)  layers  lead to different inference speedups under compiler optimization. Besides, changing the kernel size can boost model performance in some cases such as  wide activation \cite{yu2018wide,zagoruyko2016wide}, which first expands features before ReLU functions and then  uses linear low-rank convolution that factorizes a large convolution kernel into two low-rank convolution kernels without  additional parameters or computation. Thus it is desirable to find the suitable kernel size.

By transforming tiles of the input and kernel into a modulo polynomials  Winograd domain, Winograd  \cite{lavin2016fast,barabasz2019winograd}  can  reduce  the  arithmetic complexity of a CONV layer by up to a factor of 4 compared with direct convolution.  With Winograd, almost all of the arithmetic is performed by dense matrix multiplications with sufficient  dimensions, in order to improve computation efficiency, even with small batch size. 

\subsection{DNN Model Pruning}
DNNs  achieve superior performance on various practical applications such as classification or detection with  large storage/computation costs. To simultaneously reduce the storage/computation and accelerate inference speed, DNN model pruning has been proposed during DNN training for reducing the redundancy in  DNN  weights \cite{wen2016learning,guo2016dynamic,he2018amc,he2019filter}.

\paragraph{Pruning Method.} There are various pruning methods including  heuristic pruning method \cite{han2015learning,guo2016dynamic,frankle2018lottery}
such as magnitude pruning \cite{han2015learning,guo2016dynamic,liu2018rethinking}, and optimization-based pruning method \cite{zhuang2018discrimination,zhu2018ijcai,ma2019tiny} such as   Alternating Direction Methods of Multipliers (ADMM) pruning \cite{zhang2018systematic,li2019compressing}. 
The heuristic pruning is usually  performed in an iterative   manner to prune weights with small  magnitudes.
The optimization-based pruning method normally incorporates  a regularization term in  loss function  to  solve  pruning  problems. Specifically, ADMM pruning  \cite{zhang2018systematic,li2019compressing}  incorporates a dynamic regularization penalty to adjust the relative significance of the penalty and the loss, leading to more efficient training.

\paragraph{Pruning Type.} There exist a number of different pruning types for each layer, including unstructured pruning \cite{han2015learning,guo2016dynamic,liu2018rethinking}, and structured pruning \cite{min20182pfpce,zhuang2018discrimination,zhu2018ijcai,ma2019tiny,Liu2020Autocompress}. 
Unstructured pruning  removes weights at arbitrary position to significantly decrease the number of weights in DNN model. But the irregularly pruned weight matrix with indices requires additional storage and has limited hardware parallelism. 
Different from unstructured pruning, many works investigate the structured pruning methods \cite{min20182pfpce,zhu2018ijcai,Liu2020Autocompress} such as filter pruning \cite{he2019filter} 
The  highly regular model structures after pruning are  compatible with hardware parallel implementations.  But the compression ratio may be limited as pruning the whole filter can cause non-neglectable accuracy degradation. To overcome their weakness, 
pattern pruning \cite{ma2019pconv,niu2020patdnn}  combines  kernel  pattern  pruning  with  connectivity pruning. It first reserves 4 non-zero weights out of the original $3\times 3$ kernels and then  removes the whole redundant kernels and the corresponding connection between the input and output channels. Pattern pruning achieves the benefits of both unstructured and structured pruning while avoiding their weaknesses.

\subsection{Neural Architecture Search}

NAS \cite{zoph2016neural} is a popular research direction  aiming to automate the design process of DNN architectures for a given task/dataset. NAS strategies have found various neural architectures with  state-of-the-art performance outperforming human experts’ design on a variety of tasks \cite{cai2017efficient,liu2018darts,liu2018progressive,luo2018neural}. 
Similar to hyperparameter optimization, NAS can often be formulated as a black-box optimization problem \cite{elsken2018neural} and the evaluation of the objective can be very expensive due to the training of the architecture.
NAS can adopt various methods including RL \cite{zoph2016neural,liu2018progressive}, evolution \cite{wang2019evolving,fielding2018evolving} and gradients \cite{liu2018darts,cai2018proxylessnas} to design the  search strategies. RL-based NAS \cite{zoph2016neural} trains a RNN with the reinforce loss to generate different the architectures. Evolution-based NAS \cite{wang2019evolving} adopts genetic algorithms such as   population  generation  and  offspring elimination to explore different models.   Gradient-based NAS \cite{liu2018darts,cai2018proxylessnas} proposes a differentiable algorithm and solve  the problem with a bilevel optimization.

\section{Problem Motivation and Formulation}

\subsection{ Motivation }


It is desirable to achieve real-time inference execution on edge-computing platforms.
This paper mainly focuses on the 3D object detection task, which is of essential importance in autopilot for self-driving cars.
However, the existing works on 3D object detection mainly use the powerful GPUs with large memory and computation capacities for inference, which are not available on vehicles due to the cost considerations.
It is a challenging task to deploy 3D object detection with real-time inference performance on resource-limited computing devices on self-driving cars.

\begin{figure}[tb]    
 \centering
 \scalebox{1.0}{
\begin{tabular}{p{0.22\textwidth}p{0.22\textwidth}}
\includegraphics[width=0.22\textwidth]{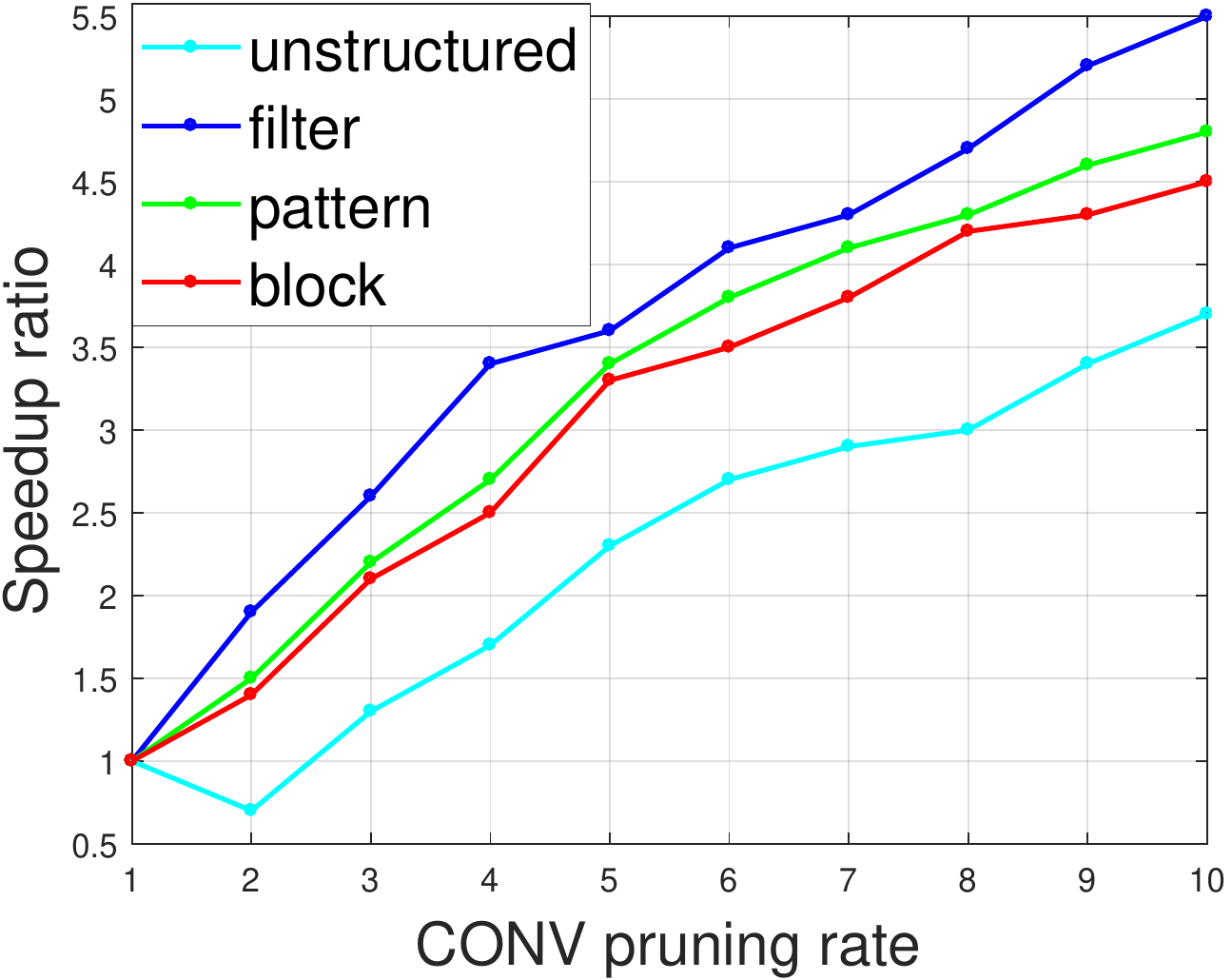}&
\includegraphics[width=0.22\textwidth]{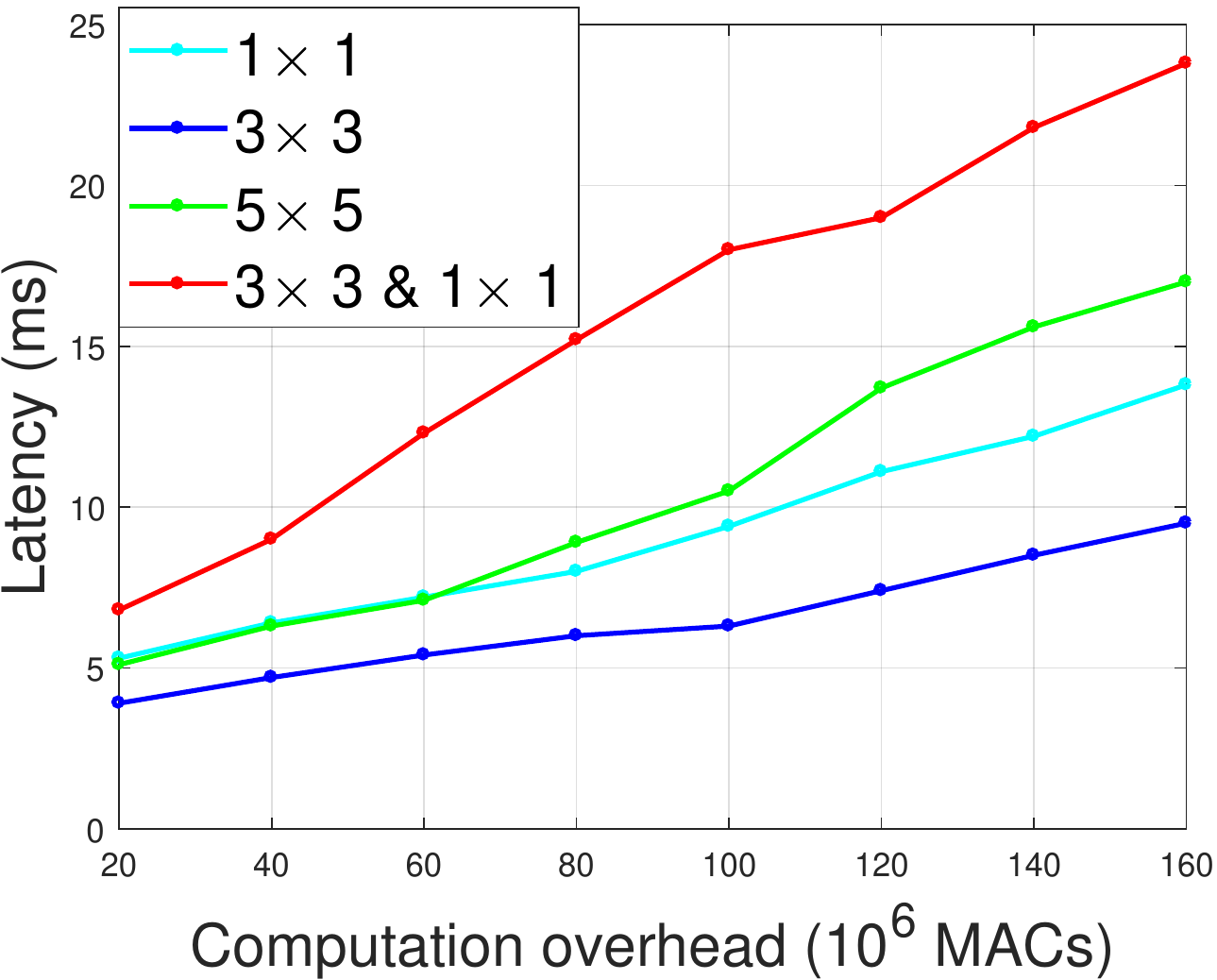} \\
\makecell{(a)} & \makecell{(b)}
\end{tabular}}
\caption{\small{(a) Speedup ratio v.s. CONV pruning rate for various pruning methods. (b)  Latency v.s. computation overhead for various kernel sizes. }} \label{fig: motivation}
\end{figure}

To satisfy the real-time requirement, we adopt model pruning  and network enhancement together with compiler optimizations to accelerate the inference and mitigate the potential  mean average precision (mAP) loss. Previous works usually have fixed pruning methods and pruning type for the whole model. 
However, we find that for the same layer, different pruning types with different pruning ratios have various inference speedup  as shown in Figure \ref{fig: motivation} (a). 
Thus, different layers may have different 
{best-suited} pruning types.   
Besides, we observe that  the layer-wise speed or latency of different kernel sizes under compiler optimization are also different as shown in Figure \ref{fig: motivation} (b). 
Moreover, Winograd can speedup the inference although incurring transformation overheads.
Thus, the performance of  network enhancement techniques needs to be re-evaluated to in terms of transformation overhead, 
inference speed and mAP. 
Meanwhile,   designing the pruning and network enhancement  requires much human expertise and efforts in hyperparameter fine-tuning.  


Based on the above observations, we propose to use a compiler-aware unified framework  with network enhancement and   pruning search to find a unified scheme so that the enhanced and pruned model can satisfy the mAP and real-time requirements for 3D object detection, thus facilitating the 3D detection deployment on edge devices. We highlight that the framework is based on RL which automatically generates the unified scheme without human assistance by searching the unified search space.

\subsection{Problem Definition}
In this problem, we  search a unified scheme with network enhancement and pruning,
such that the enhanced and pruned model can keep the  mean average precision (mAP)  as high as possible while  satisfying the latency or real-time requirement, that is, the whole inference time should not exceed a threshold. The problem can be formulated as below,
{\small
\begin{align} \label{eq: pro}
	\min_s \quad & \mathcal{L}_{val}( \hat\theta(s)), \\
	\text{s.t.} \quad & \hat\theta(s) = \mathrm{argmin}_\theta \enskip \mathcal{L}_{train}(\theta(s)), \label{eq: map_require} \\   & t(\hat\theta(s)) \leq T \label{eq: time_require},
\end{align}}%
where $s$ denotes a unified scheme including a list of enhancement and pruning actions, and $\theta(s)$ denotes the  new model parameters  following the unified scheme. $ \mathcal{L}_{train}$ and $ \mathcal{L}_{val}$ represent the training and the validation loss, respectively. $t(\cdot)$ denotes the inference time of the model and $T$ is the real-time or latency requirement for inference. The constraint \eqref{eq: map_require} is the retraining or fine-tuning step after performing the unified scheme to further improve mAP and mitigate the mAP loss. The constraint \eqref{eq: time_require} is the real-time requirement that its inference time after enhancement and pruning should not go beyond a threshold.

\subsection{Unified Search Space}

For solving problem \eqref{eq: pro}, we search a unified space to find a unified scheme incorporating the network enhancement and model pruning. 
For network enhancements,  we employ intra-kernel replacement  to choose a suitable kernel size and explore the usage of 
Winograd \cite{lavin2016fast} for inference acceleration.  For pruning search, we  choose the pruning method, pruning type and pruning ratio separately.  We  list the choice of the network enhancement and pruning search   in Table \ref{Table: search_space} and specify them in the following sections.

\begin{table}[tbh]
\caption{Unified search space}
\begin{center}
\begin{threeparttable}
\scalebox{0.99}{
\begin{tabular}{c |c | c }
    \toprule
         \multirow{2}{*}{ \makecell{ Network  \\ enhancement}}   & \makecell{  Intra-kernel \\ replacement } & \{$1\times 1$, $3\times 3$,  DW $3\times 3$  \& $1\times 1$ \}\\ 
               \cline{2-3}
               & Winograd   & \{0, 1\} \tnote{a} \\
               \midrule
        \multirow{7}{*}{ \makecell{ Pruning \\ search}}   & \multirow{2}{*}{ \makecell{ Pruning \\ method}}    &      Magnitude pruning \cite{frankle2018lottery,liu2018rethinking}                \\
    &  & ADMM  pruning \cite{zhang2018systematic,li2019compressing}  \\
      \cline{2-3}
  &   \multirow{3}{*}{\makecell{ Pruning \\ type}}     &       Filter pruning \cite{he2019filter}               \\
  &   &  Pattern pruning \cite{ma2019pconv} \\
  &    &  Block  pruning  \\
      \cline{2-3}
  &   \makecell{  Pruning  \\ ratio}   &    \{ 0, 0.3, 0.5, 0.7, 0.8, 0.9\}                  \\ 
      \bottomrule
\end{tabular}}
\begin{tablenotes}
\item[a] \small  1 and 0 denote using  this method or not; 
\end{tablenotes}
\end{threeparttable}
\end{center}
\label{Table: search_space}
\end{table}

\subsubsection{Network Enhancement Space}
The network enhancements include  intra-kernel replacement  to choose a suitable kernel size and {whether to use} Winograd \cite{lavin2016fast,barabasz2019winograd} for inference acceleration. 

\paragraph{Intra-kernel replacement.}  Under compiler optimization, different kernel sizes result in various speedups. Besides, changing kernel size may boost model performance \cite{yu2018wide}. Thus we consider to search different kernel sizes. As the compiler can achieve higher accelerations on small kernel sizes, we mainly choose between $1\times 1$, $3\times 3$, and a cascade of depthwise (DW)  $3\times 3$  \& $1\times 1$ \cite{howard2017mobilenets,sandler2018mobilenetv2}.

\paragraph{Winograd.} Winograd \cite{lavin2016fast,barabasz2019winograd}  can reduce the arithmetic complexity of a CONV layer by
up to a factor of 4 compared to direct convolution. However, the transformations in Winograd cause certain {computational} overheads which could degrade the performance improvement.
Thus, we explore the usage of Winograd to confirm whether it can improve the inference speed. 

\subsubsection{Pruning Search Space}
The pruning scheme includes three aspects: pruning method, layer-wise pruning type and pruning ratio. 
For pruning methods, we search from magnitude pruning \cite{frankle2018lottery} and ADMM pruning \cite{zhang2018systematic}. {Different pruning methods could lead to different pruning locations even with the same pruning type and ratio.}
For pruning types, we explore the space of filter pruning \cite{he2019filter}, pattern pruning \cite{ma2019pconv} and block pruning (specified below)  for each layer.
Besides, in each layer, along with the pruning type, we choose a pruning ratio from $ \{ 0, 0.3, 0.5, 0.7, 0.8, 0.9,  \}$, denoting the percentage of pruned weights. 0 means skip pruning. 

Note that in previous pruning \cite{zhang2018systematic,he2019filter,ma2019pconv,guo2016dynamic},  all the DNN layers usually share the same  pruning type  
under a fixed pruning method, {and the layer-wise pruning ratio is manually predefined.}
However, each layer may have different {best-suited} 
pruning types under the compiler optimization {due to its unique computation pattern and  layer size}. Thus, different  from the fixed pruning scheme, we use configurable pruning design, which allows each layer to choose its own pruning type and its corresponding pruning ratio, and  the whole model to choose a pruning method.

\paragraph{Block pruning.} As pattern pruning \cite{ma2019pconv} can only be applied to 3×3 CONV  layers,  we  propose block pruning as a more general fine-grained structured pruning type compatible with various CONV kernel sizes. More specifically, the weights of a CONV layer has the shape $c_{in} \times c_{out} \times n \times n $ where $c_{in}$ and $ c_{out}$ are the input and output channel numbers respectively for this layer,  and $n$ is the kernel size. We first divide the weights into $b_{in} \times b_{out}$ blocks, and each block has the shape $ \frac{c_{in}}{b_{in}} \times \frac{c_{out}}{b_{out}} \times n \times n $. So the block division  is carried out on the input and output channels instead of the kernels. Next, in block pruning, all of the kernels in the same block  share the same pruning patterns, which means we prune the same locations for all kernels in this block.  The pruning patterns of different blocks can be different, so the pruned locations of two kernels from two different  blocks are not necessarily the same. With block pruning, we can  leverage hardware parallelism  with compiler optimizations to improve inference speed.

\subsection{Compiler Assistance} \label{sec: compiler}

To achieve real-time performance on edge devices, we incorporate compiler optimization to accelerate inference. 
{Therefore, we should be able to obtain the inference speed performance of pruned and enhanced models under the given compiler optimizations.}
More specifically, we build up an automatic code generation framework to incorporate compiler optimizations. 
The compiler optimizations consist of several components including domain  specific  language related optimizations, sparse model storage, matrix reorder, etc.  \cite{ma2019pconv}  to improve the inference speed on edge devices. We show more details in Appendix \ref{app: compiler detail}. 
Given a pruned and enhanced model, the framework can dynamically perform compiler-level optimizations to accelerate the inference of each layer in the model  based on the pruning type and pruning ratio.   The framework supports various pruning types such as filter pruning \cite{he2019filter}, pattern pruning \cite{ma2019pconv} and { block pruning}.  We demonstrate  the superior inference acceleration performance  on  sparse DNN models in Appendix \ref{app: compiler performance}.
Thus, the framework can provide the speed performance of DNN models with compiler optimizations.

\section{Unified Framework with RL}

\begin{figure}[t]  
\centering  
\includegraphics[width=0.95\columnwidth]{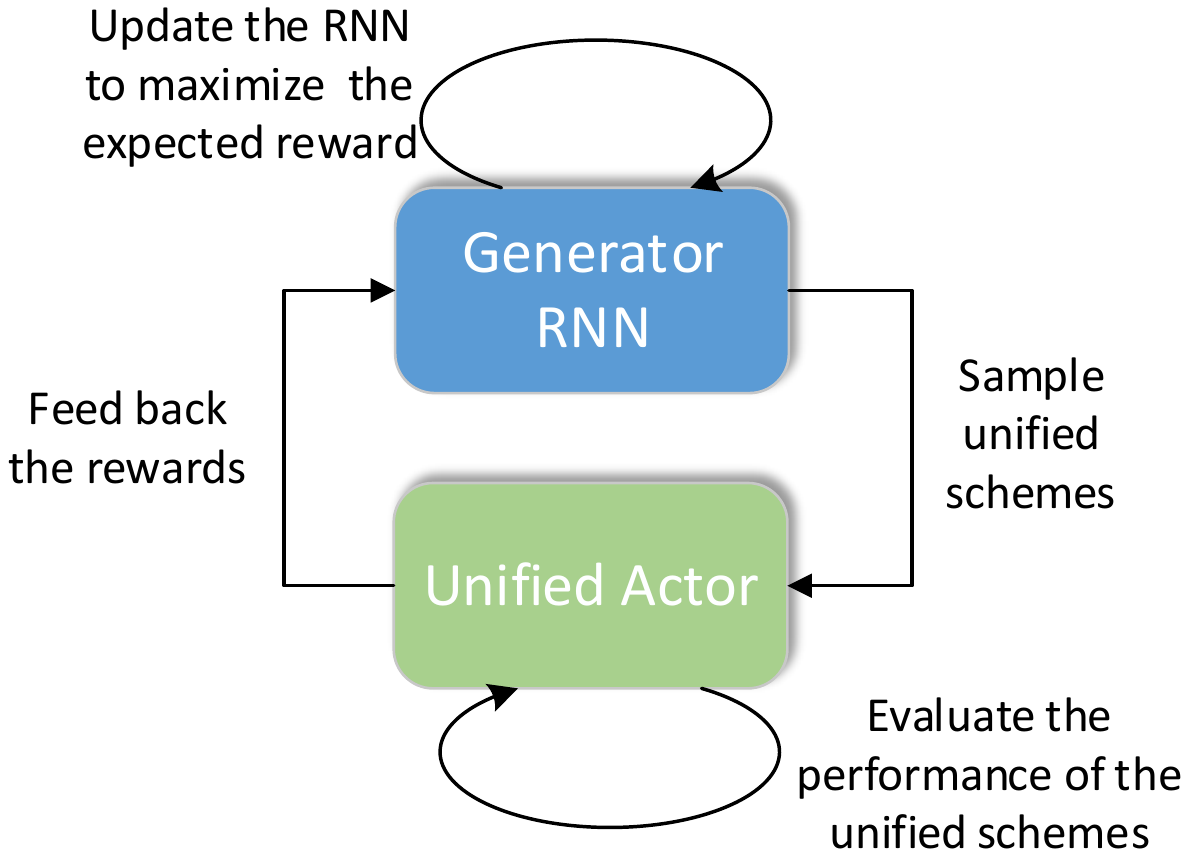}  
\caption{Unified network enhancement and  pruning search framework.}  
\label{Fig: framework} 
\end{figure}

In this  section, we  first describe the framework of the proposed unified network enhancement and pruning search as shown in Figure \ref{Fig: framework}. It has two main functional components: the generator RNN and the unified actor.  The generator is implemented by a RNN to generate various unified schemes. After receiving the unified scheme, the unified actor starts to perform the network enhancement and  pruning on the original model following the unified scheme. Once finished, its reward (the mAP and speed performance)  is sent back to the generator RNN. Then the generator RNN is updated with a policy gradient method  to maximize the expected reward of the sampled schemes.

\subsection{ Setup}

Before the generator RNN outputs the unified schemes, we need to initialize the generator RNN and perform some preliminary analysis  for the given  model. In the 3D object detection with point clouds, we train a PointPillars model \cite{lang2019pointpillars} as the original starting model. Note that we do not choose other 3D sparse convolution based models such as PV-RCNN \cite{shi2020pv} or SECOND \cite{yan2018second} since the PointPillars model has the best inference speed performance with competitive mAP. For example, when executed on Nvidia high-end GPU, it has inference latency of 20ms, which is at least 2.7 times faster than other models. Besides, other methods are not  well supported by compiler optimization.
Given the PointPillars model, we first investigate the model architecture and obtain the statistics of the whole model and each layer, including its parameter shapes,  number of multiply–accumulate  operation (MAC) and so on. This information can help with the compiler optimizations.

\subsection{Generator RNN}

\subsubsection{Unified Scheme Generation}

\begin{figure}[t]  
\centering  
\includegraphics[width=0.95\columnwidth]{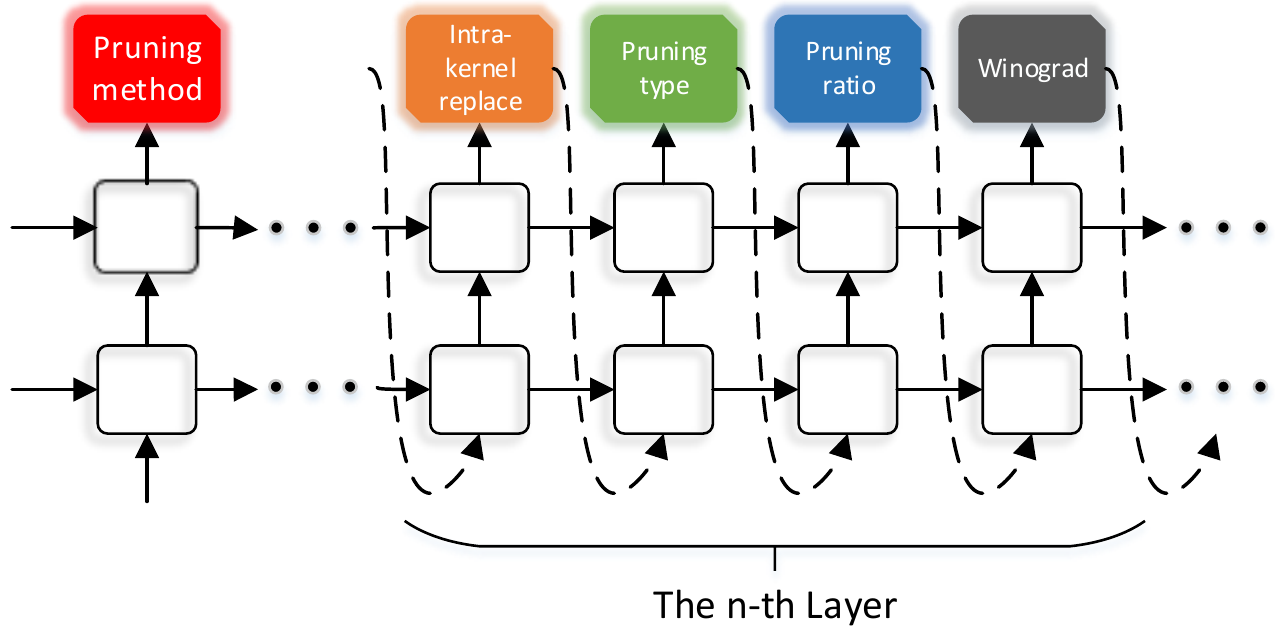}  
\caption{Unified scheme generation.}  
\label{Fig: rnn} 
\end{figure}

In the unified framework, a generator RNN is applied to generate the unified schemes. 
The generator RNN generates the details of a unified scheme including network enhancement options (intra-kernel replacement and Winograd for each layer) and pruning search exploration (pruning method,  layer-wise pruning type and its corresponding  pruning ratio), as a sequence of tokens shown in Figure \ref{Fig: rnn}. Every prediction is carried out with  a softmax classifier and then fed into the next time step as input.
Once the generator RNN finishes generating a unified scheme, it will be evaluated and its reward is used to train the parameters of the generator RNN, $\theta_c$.

\subsubsection{Generator Training}

The list of tokens that the  generator  RNN predicts represents a list of actions $s_{1:Q}$ for a unified scheme, where $Q$ is the number of total  hyperparameters or actions. Following this unified scheme, the processed model achieves a  mAP $A$ on the test dataset with an average inference time $t$ on edge platforms.  We define the reward function as the following,
{\small
\begin{align} \label{eq: reward}
    R = A - \alpha \cdot \mathrm{max}(0, t - T),
\end{align}}%
where $T$ is the real-time requirement for inference and $\alpha$ controls the relative significance of the mAP and inference time. Basically, if the processed model achieves high mAP and its inference time does not exceed the threshold, it will get a high reward. Otherwise if $A$ is small or $t>T$, the reward becomes smaller. 


The generator is trained to maximize its expected reward which is  non-differentiable. To deal with this, we adopt   an empirical approximation of the  REINFORCE rule \cite{williams1992simple} and its gradients can be given by the following,
{\small
\begin{align} \label{eq: gradients}
 \bigtriangledown_{\theta_c} J(\theta_c)  = \frac{1}{K} \sum_{k=1}^{K} \sum_{q=1}^{Q} \bigtriangledown_{\theta_c} \log P(s_q|s_{(q-1):1};\theta_c)(R_{k} - b),
\end{align}}%
where $P(s_q|s_{(q-1):1};\theta_c)$ is the probability of the action $s_q$ in the unified scheme $s$ from the generator RNN $\theta_c$. 
$K$ is the number of different unified schemes  sampled by the generator in one batch.
The reward of  the $k$-th unified scheme   is $R_k$.  
We introduce $b$ as the exponential moving average of the previous reward to reduce the estimate variance. We show the reinforcement training progress in Algorithm \ref{alg: framework}.

\begin{algorithm}[t]
\small
  \KwInput{Original model;}
  \KwOutput{The best unified scheme and pruned model;}
  Setup  to get statistics of the original model;\\
  Initialize the parameters of GR$^1$ ; \\
  \algorithmfootnote{$^1$ GR denotes the generator RNN. \\ $^2$  UA denotes the unified actor.} 
  \For{each step}    
        { 
        	GR samples $K$ unified schemes; \\
        	UA selects $B$ unified schemes with the BO acquisition function from the $K$ schemes; \\
        	  \For{ each selected scheme} 
        	  { 	UA   performs evaluation to obtain the mAP and speed performance; \\
        	  }
        	UA sends  the evaluation performance to GR;\\
        	GR  obtains the gradients with rewards; \\
        	GR updates its parameters; \\
        	UA updates the GP of BO with the new observation data;\\
        }
    Fine-tune the best pruned model to further improve its performance; \\ 
\caption{Unified framework} \label{alg: framework}
\end{algorithm}

\subsection{Unified Actor with Bayesian Optimization}

The main function for the unified actor is to obtain the mAP and speed performance of the $K$ unified schemes from  the generator. However, evaluating  the scheme performance  is expensive as it needs pruning and training the model with certain epochs. Moreover, the search space is  large, leading to difficulties for finding a scheme with a good performance. Thus we first use Bayesian optimization (BO) \cite{snoek2012practical,klein2017fast,chen2018bayesian} with Weisfeiler-Lehman  (WL) kernel \cite{morris2017glocalized,shervashidze2011weisfeiler} to  select a subset of unified schemes from the generator. Then we perform actual evaluation for the selected schemes.

\subsubsection{BO with WL Kernel}

To accelerate the training process, we propose to use BO to help with the unified actor for reducing the number of actual evaluated schemes. 
More specifically, given $K$ unified schemes, before we actually evaluate them, BO is first utilized to find $B$ ($B<K$) unified schemes through the acquisition function. The selected $B$ schemes may achieve better performance with high probabilities while the rest $K-B$ schemes are potentially weak. Thus, during  evaluation, we only evaluate the  selected $B$ unified schemes and obtain their real performance while the rewards of the rest  $K-B$ schemes are obtained with the Gaussian process (GP) surrogate  model. 
However, conventional BO cannot be directly employed  since  the unified schemes can be treated as  a  directed acyclic  graph (DAG). 
To facilitate this, we  use a GP with a WL graph kernel  which can naturally deal with the non-continuous graphs. 

The WL kernel compares two directed graphs   as below, 
{\small
\begin{equation}
    k^{H}_{\mathrm{WL}}(s, s') = \sum_{h=0}^H w_h k_{\mathrm{base}}\bigl(\phi_{h}(s), \phi_{h}(s')\bigr),
    \label{eq:wl}
\end{equation}}%
where $h$ denotes the  WL iteration index and $H$ denotes the maximum iteration level. In the $h$-th WL iteration, it first collects the  graph features $\phi_{m}(s)$ and $\phi_{m}(s')$ for two graphs $s$ and $s'$, and then compares the two graphs with  $k_{\mathrm{base}}\bigl(\phi_{m}(s), \phi_{m}(s')\bigr)$. The final WL kernel is  the weighted sum of the comparison in each iteration. We employ  the dot product as the  base kernel $k_{\mathrm{base}}$ and  $w_h$ are the weights of each WL iteration, which are set to equal following \cite{shervashidze2011weisfeiler}. The details of  WL kernel are demonstrated in Appendix \ref{app: klkernel}. We adopt the expected improvement \cite{qin2017improving,frazier1807tutorial}  as the acquisition function in the work.

\subsubsection{Scheme Evaluation}

In the scheme evaluation, for each unified scheme, the unified actor processes the original model following the unified scheme. 
It first perform the intra-kernel replacement by replacing the kernel with new kernel sizes and finetuning the new model with a few epochs. 
Then it starts to prune the model.
It can perform multiple kinds of pruning method such as ADMM pruning \cite{zhang2018systematic} or magnitude pruning \cite{frankle2018lottery} with supports for various pruning types and pruning ratios.

After pruning and retraining, we can obtain the mAP of the pruned model. The inference speed of the pruned model can be obtained by the compiler optimization framework with the configuration of Winograd from the scheme as shown in Section \ref{sec: compiler}. The compiler optimization can be performed in parallel with the finetuning step in the pruning during evaluation.

\subsection{Acceleration Techniques} \label{sec: acceleration}
The unified framework requires to prune the model multiple times  with  different pruning schemes to train the generator RNN until convergence.  It could be time-consuming to collect the large pruning performance data. To accelerate the generator training, we employ several acceleration techniques including parallel updating and early stop as demonstrated in Appendix \ref{app: accelerate} with details.

\section{Experimental Results}


\subsection{Experiment setting}

All experiments are conducted with the KITTI object detection benchmark dataset \cite{Geiger2012CVPR}, which consists of samples with both lidar point clouds and images. We follow the standard convention  \cite{zhou2018voxelnet} of only using lidar points,  but compare with fusion methods that use both lidar and images. We train a PointPillars  for cars  use a pillar resolution of 0.24 m with   12000 as the max number of pillars  and 100 as
the  max number of points per pillar. More results for  pedestrians and cyclists are shown in Appendix \ref{app: pedestrian}. 
During inference, we apply axis aligned non maximum suppression (NMS) with an overlap threshold of 0.7 intersection-over-union (IoU) following the official KITTI  protocol.

The generator RNN is a two-layer LSTM with 49 hidden units on each layer. Its weights  are initialized uniformly between -0.1 and 0.1. The ADAM optimizer \cite{kingma2019method} is employed with a learning rate of 0.0005. 
For the parallel training,  we  use 50 GPUs to concurrently evaluate the unified schemes. The generator training takes about 10 days. During the evaluation, for  each unified scheme, we first perform intra-kernel replacement and fine-tuning the model with 5 epochs. Then we start to prune the model. For ADMM pruning, we prune with 5 epochs and fine-tune the pruned model with 10 epochs. For magnitude pruning, we first prune the model according to weight magnitudes and then fine-tune the model with 10 epochs. 
Note that we start from a well-trained model and thus it does not need too many epochs to obtain competitive results with network enhancement and pruning. We use the ADAM Optimizer \cite{kingma2019method}
with a learning rate of 0.0002, weight decay of 1e-4 and momentum of 0.8. More training hyperparameter settings are shown in Appendix \ref{app: exper}.

\subsection{Comparison with the Original Model }

\begin{table}[tb]
\caption{Comparison with the original model}
\begin{center}
\begin{threeparttable}
\scalebox{0.97}{
\begin{tabular}{c|c|c|c|c|c}
\toprule
\multicolumn{2}{c|}{}                                                                      & \begin{tabular}[c]{@{}l@{}}PointPillars\\ (0.16)\end{tabular} & \begin{tabular}[c]{@{}l@{}}Ours\\ (0.16)\end{tabular} & \begin{tabular}[c]{@{}l@{}}PointPillars\\ (0.24)\end{tabular} & \begin{tabular}[c]{@{}l@{}}Ours\\ (0.24)\end{tabular} \\ \hline
\multicolumn{2}{c|}{Parameter \#}                                                          & 5.8M                                                               & 1.1M                                                       & 5.8M                                                               & 0.8M                                                       \\  
\multicolumn{2}{c|}{ Computation  \# (MACs) }                                                 & 60G                                                                & 10.7G                                                       & 28G                                                                & 3.9G                                                       \\  \hline
\multicolumn{2}{c|}{Server GPU speed (ms)}                                                        & 25                                                                 & 24                                                         & 20                                                                 & 18                                                         \\  
\multicolumn{2}{c|}{Mobile GPU speed (ms)}    &    542  &  189 &  257    &  97  \\ \hline
\multirow{3}{*}{ \makecell{ Car  3d \\ detection \\ AP} } & Easy   
                   &    84.69   &  85.50  &    84.05  &     85.20    \\
        & Moderate &  75.11  &   76.58     &  74.99   &  75.57          \\
        & Hard  &   69.53  &    70.58    &    68.30   &     68.37       \\ \bottomrule
\end{tabular}
}
\end{threeparttable}
\end{center}
\label{tab: original}
\end{table}

\begin{table*}[tbh]
\center
\caption{Comparison of 3D detection methods}
\begin{center}
\begin{threeparttable}
\scalebox{1.1}{
\begin{tabular}{c|c|c|c|c|c|c|c|c|c}
\toprule
\multirow{2}{*}{ Detection methods} & \multirow{2}{*}{Modality} & \multirow{2}{*}{ \makecell{ Server GPU\\ speed (ms)}} & \multirow{2}{*}{ \makecell{ Mobile GPU\\ speed (ms)}}  & \multicolumn{3}{c|}{Car 3D detection} & \multicolumn{3}{c}{Car BEV detection} \\
\cline{5-10}
                         &                           &       &                 & Easy      & Moderate      & Hard     & Easy      & Moderate      & Hard      \\
\midrule

           F-PointNet \cite{qi2018frustum}       &      \multirow{3}{*}{R+L\tnote{a} }                   &    170    &   -       &    82.13               &   69.22 & 60.78 & 90.58 & 84.73 & 75.12   \\
           AVOD-FPN \cite{ku2018joint}  &           & 100      &   - &        82.77 & 71.94  & 66.31 & 90.64 & 84.37  & 80.04.       \\
           UberATG-MMF \cite{liang2019multi}   &       &      80    &-  &       86.81 & 76.75 &  68.41    &   89.49 &  87.47  & 79.10           \\
           \midrule
           Fast Point R-CNN   \cite{chen2019fast}    &      \multirow{5}{*}{L }      &       65   & -  &    85.39 & 77.46 & 70.21 & 89.97 & 87.08 & 80.40      \\
         STD \cite{yang2019std}    &   &    80 &  - &  86.61  & 77.63 & \textbf{76.06} &  89.66 &  87.76 &  \textbf{86.89}   \\
           SECOND \cite{yan2018second}  &    &   50 & - &     83.34 &  72.55  &  65.82 &  89.39  & 83.77  &  78.59   \\
           Point-GNN \cite{shi2020point} &    &    643  & -  & \textbf{87.25} & \textbf{78.34} & 72.29 & \textbf{92.04} & \textbf{88.20} & 81.97   \\
           Ours     &     &  \textbf{ 18 }  &   97   &   85.20 & 75.57 & 68.37   & 90.02 & 86.79 &  80.80   \\
\bottomrule
\end{tabular}}
 \begin{tablenotes}
 \item[a] \small  'L' and 'R' represent  LiDAR and RGB images respectively; \\
 \end{tablenotes}
\end{threeparttable}
\end{center}
\label{tab: 3ddetection}
\end{table*}

The original unpruned model is based on PointPillars \cite{shi2020point}. We compare the performance of the original model and the model found by our method shown in Table \ref{tab: original}. We try   two pillar grid sizes $0.16 \times 0.16$ and $0.24 \times 0.24 $ (large grid sizes means small input size for the model) and their real-time requirement  are set to 200ms and 100ms respectively. 
We can make the following observations.
(1) Although the parameter numbers are not large, the computation numbers (MACs) are relatively large due to the tremendous CONV operations, leading to difficulties for achieving real-time on mobile. 
(2) Small input size results in obvious computation reduction (60G v.s. 28G for original PointPillars model). 
(3) As the server GPU  (GTX 1080Ti) is far more powerful than the mobile  (Samsung Galaxy S20 phone) GPU, the speed on mobile is much slower than that on server GPU (e.g. 542ms v.s. 25ms), thus it is more challenging to achieve real-time on mobile.
(4) Although the parameter and computation number are reduced significantly by pruning, the speed on powerful server GPUs does not naturally get obvious improvements (e.g. 25ms v.s. 24 ms) without further optimizations. 
(5) Different from the server speed, with compiler optimization, the mobile speed can be improved significantly  after pruning (with about 2.7$\times$ speedup).    
(6) The detection performance of the our models are better that the original model indicating that there may be certain overfitting or redundancy in the  original model, resulting in the lower performance.

Based on the above observations, we mainly focus on the configuration with grid size  $0.24 \times 0.24 $. We demonstrate that the proposed method can achieve real-time inference (97ms)  on mobile devices (Samsung Galaxy S20 phone)  with state-of-the-art detection performance. Although there are some other DNN inference  acceleration  frameworks such as Tensorflow-Lite \cite{TensorFlow-Lite}, TVM \cite{chen2018tvm} and MNN \cite{Ali-MNN}, we are the first to support real-time 3D object detection on mobile GPUs.

\subsection{Comparison of 3D Detection Methods }

We show the performance of  the model obtained through the proposed method compared with state-of-the-art 3D object detection methods as shown in Table \ref{tab: 3ddetection}. We obtain the statistics of other methods by following their released code or model with the default setting. 
The server GPU inference time is obtained on  GTX 1080Ti and the mobile GPU speed is based on the Samsung Galaxy S20 phone.

We can observe that the proposed method can achieve competitive 3D detection performance with the fastest speed.
On server GPU, it only needs 18ms to process one input sample during  inference on average.  
Compared with other methods not based on PointPillars, the speed are at least 2.7$\times$ faster with competitive detection performance.
On mobile GPU, our method achieves an average 97ms inference speed, satisfying the real-time requirement, while we  are not able to perform compiler optimization and obtain their mobile speed  for most of the other 3D object detection methods.
The reason is that their proposed various special layers or structures  to deal with the sparse data  are currently not supported by the compiler optimization. For example, the 3D sparse convolutions \cite{graham2015sparse} which are adopted by  many 3D detection works uses highly irregular data and computation patterns such as Hash table, leading to difficulties of compiler optimizations to improve computation parallelism. 
 


\subsection{Comparison of Pruning Frameworks}

\begin{table}[tb]
\caption{Comparison of various pruning methods}
\begin{center}
\begin{threeparttable}
\scalebox{0.91}{
\begin{tabular}{c|c|c|c|c|c}
\toprule
\multirow{2}{*}{ \makecell{Pruning \\ methods}}  & \multirow{2}{*}{\makecell{ server GPU\\ Speed  (ms)}}  & \multirow{2}{*}{\makecell{ mobile GPU \\ Speed  (ms)}}  & \multicolumn{3}{c}{Car 3D detection} \\
\cline{4-6}
            &    &  & Easy      & Moderate      & Hard        \\
\midrule
          Original   &  20 & 257  & 84.05  & 74.99  & 68.30 \\
           Filter \cite{he2019filter}  &  18 & 81 & 81.54 &  68.10 &  65.90  \\
           Pattern \cite{ma2019pconv} &  19 & 111 &   80.97 &  73.77 & 68.05         \\
            Block       &  20 & 143 &  80.88 & 72.98 &  67.85         \\
           Ours     & \textbf{  18 }  & 97 &  \textbf{ 85.20} & \textbf{ 75.57} & \textbf{ 68.37 }   \\
\bottomrule
\end{tabular}}
\end{threeparttable}
\end{center}
\label{tab: pruning}
\end{table}

We compare the performance of the proposed method with other pruning types under the ADMM pruning framework \cite{zhang2018systematic} as shown in Table \ref{tab: pruning}. All of these methods start from the same PointPillars model. Their pruning ratio are set to the same with the overall pruning ratio of our pruned model (86\%).
As observed, the proposed method can achieve the best detection performance compared with other methods where all the layers share the same pruning type, demonstrating the advantages of using flexible pruning types for each layer. 
Besides, with compiler optimization, filter pruning is the fastest but suffers from obvious detection performance degradation. The proposed method can process one lidar image within  97ms on average  with the highest precision, demonstrating the superior performance of the proposed method to achieve real-time inference on mobile with state-of-the-art detection performance. 





\subsection{Ablation Study}

We explore the effects of various pillar grid sizes and show the performance in Appendix \ref{app: ablation}. Besides, we test the case with pruning search only instead of unifying network enhancement and pruning search. Under the same configuration, pruning search only can achieve a mAP of 74.89\% with an average inference time of 108ms. Compared with the unified framework with 76.38\% mAP and 97ms inference, we can see that incorporating network enhancement can further improve the detection and speed performance. Moreover, As we increase the real-time requirement from 100ms to 90ms, the proposed method achieves a mAP of 74.15\% with an average 89ms speed, demonstrating that the method prune the model harder to improve speed, thus incurring detection performance degradation. 
More details are shown in Appendix \ref{app: ablation}.

\section{Conclusion}
We propose a unified framework incorporating network enhancement and pruning search with reinforcement learning to achieve real-time inference on mobile for 3D object detection.  The compiler-aware framework includes the detection and speed performance during the optimization and employ BO to accelerate the training. The  experimental results show that it can satisfy the real-time requirement on mobile with competitive detection performance compared with state-of-the-art 3D detection works.

{\small
\bibliographystyle{ieee_fullname}
\bibliography{egbib}
}

\clearpage

\section*{Appendix}
\setcounter{equation}{0}
\setcounter{figure}{0}
\setcounter{table}{0}
\makeatletter
\renewcommand{\theequation}{A\arabic{equation}}
\renewcommand{\thefigure}{A\arabic{figure}}
\renewcommand{\thetable}{A\arabic{table}}
\appendices

\section{Compiler Optimization Details} \label{app: compiler detail}
Compiler optimization can support all kinds of pruning types and pruning ratios. 
It can support filter pruning and pattern pruning originally.  For block pruning,
as all filters in each block share the same pruning patterns, the compiler can skip accessing the same input data corresponding to the pruned weights, thus reducing the memory access pressure among these filters. Besides the memory access, for the computation, since the pruned locations are the same among the filters in each block, they have the same computation indices, which is beneficial for computation parallelism without computation divergence within each block. So block pruning can be implemented more efficiently for computation-intensive CONV layers.

The compiler optimization consists of the following components in details.

 \paragraph{DSL related optimization.} 
 There are multiple kinds of operators with various computation patterns in the DNN models. We use a new domain specific language (DSL) to represent each DNN model. In this DSL, each layer is represented by a new layer-wised representation (LR). This DSL can show the computation graph within the model. Based on this DSL,  we can further utilize some computational graph transformation optimizations. For example,  a combination of Convolution layer/Depthwise Convolution layer and  BatchNorm layer can be fused into one layer to reduce the data movement and access, thus increasing the  instruction level parallelism.

Layer fusion is a common technique in DSL optimization to  fuse the computation operators in computation graph.  With layer fusion, we can reduce the operator number, and avoid saving the parameters of fused operators and their intermediate computation results.
We identify the operators which can be fused based on the computation laws such as associative property and distributive property, and the data access patterns.
But as the number of operator combinations is huge in the computation graph, we constrain the search of the potential operator fusion by considering whether the fusion can enlarge the overall computation for CPU/GPU utilization improvement and reduce the memory access for memory efficiency.

 \paragraph{Sparse model storage.}
 To further improve data locality, we use a more compact format to store  the sparse weights of the  model compared with   the well-known CSR. Basically, we try our best to  avoid storing zero-weights in the model with a high compression rate. This is achieved  though  removing redundant indices from the structured pruning. The sparse model storage can save the scarce memory-bandwidth of mobile devices.

 \paragraph{Matrix reorder.} 
Although structured pruning can transform model kernel matrices into small blocks with various pruning patterns, there are still some well-know challenges for sparse matrix multiplications, such as the heavy load imbalance among each thread, and irregular memory accesses. To deal with these challenges, we adopt a matrix reorder method by leveraging the structure information from the structured  pruning. Take the column pruning as example where the whole columns are pruned.  As the whole columns are removed, a certain degree of regularity appears since the rest weights are stored in unpruned columns. Based on this, matrix reorder first  rearranges the rows with the same or similar patterns together, i.e.,  reorders the rows.  
Then, matrix reorder makes the   weights in the column direction (e.g., kernels in CNN) more compact.

 \paragraph{Parameter auto-tuning.} 
During the compiler optimization, there are many parameters related to the compuation, such as data placement on GPU memory,  loop unrolling  factors, matrix tiling sizes, etc. To find the best configuration of the parameters, the compiler adopts an auto-tuning process similar to  other acceleration frameworks such as TVM \cite{chen2018tvm}. More specifically, a genetic algorithm is employed to explore the parameter space. Besides, the explore efficiency can be improved by increasing the population number in each generation to improve the  exploration parallelism.

\section{Compiler Optimization Performance} \label{app: compiler performance}

We demonstrate the efficacy of  compiler optimization   through three interesting and important DNN applications, style transfer, DNN coloring, and super resolution. 
The style transfer model is trained on Microsoft COCO dataset and employs  a generative network.  DNN coloring can convert gray images into colorful images.  It  trains on the Places scene dataset to  obtain  a novel architecture that can jointly extract  and  fuse global and local features  to perform the final colorization. 
The super resolution can convert low resolution images into images with high resolutions.  It is trained on the DIV2K dataset and mainly uses residual blocks with wider activation and linear low-rank convolution~\cite{yu2018wide}. 
With structured pruning and compiler optimization, we implement the models on a Samsung Galaxy S10 mobile phone. 
We show the  average inference time of three applications  on  mobile device in Table~\ref{table_inference}. 
We can observe that for  the inference speed, compared with the unpruned model,  structured pruning and compiler optimization can achieve    speedups of $4.2\times$, $3.6\times$, and $3.7\times$ for style transfer, coloring and super resolution, respectively. 
Without compiler optimization, the pruning model can be faster than the unpruned model. However, their inference time are still a bit longer than the real-time requirement. For example, the super resolution takes about 192ms without compiler optimization.  
With compiler optimization, we can observe that their inference speeds are further improved with a ratio of at least $2\times$. It largely reduces the inference time (e.g., from 178ms to 67ms for style transfer), satisfying the real-time requirement.
We note that all inference can complete within 75 ms, showing the possibility of achieving real-time executions of complex DNN applications on mobile.
We demonstrate that compiler optimization can further improve the speed with significant acceleration performance.

\begin{table}[tb]
 \centering
  \scalebox{0.92}{
   \begin{threeparttable}
\begin{tabular}{c|c|c|c}
\toprule[1pt]
 Inference time (ms) & \makecell{Style} &  \makecell{coloring}   &  \makecell{Super resolution} \\ 
\midrule[1pt]
 Unpruned  & 283 &  137 &  269 \\ 
\hline
Pruning  & 178  &  85 & 192 \\ 
\hline
\makecell{ Pruning + compiler } & 67 &  38 & 73 \\ 
\bottomrule[1pt]
 \end{tabular}
\end{threeparttable}}  
  \caption{Average Inference Time on the Mobile Device}
  \label{table_inference}
\end{table}

\section{WL Kernel} \label{app: klkernel}
\begin{figure}[t]
    \centering
    \includegraphics[width=0.48\textwidth]{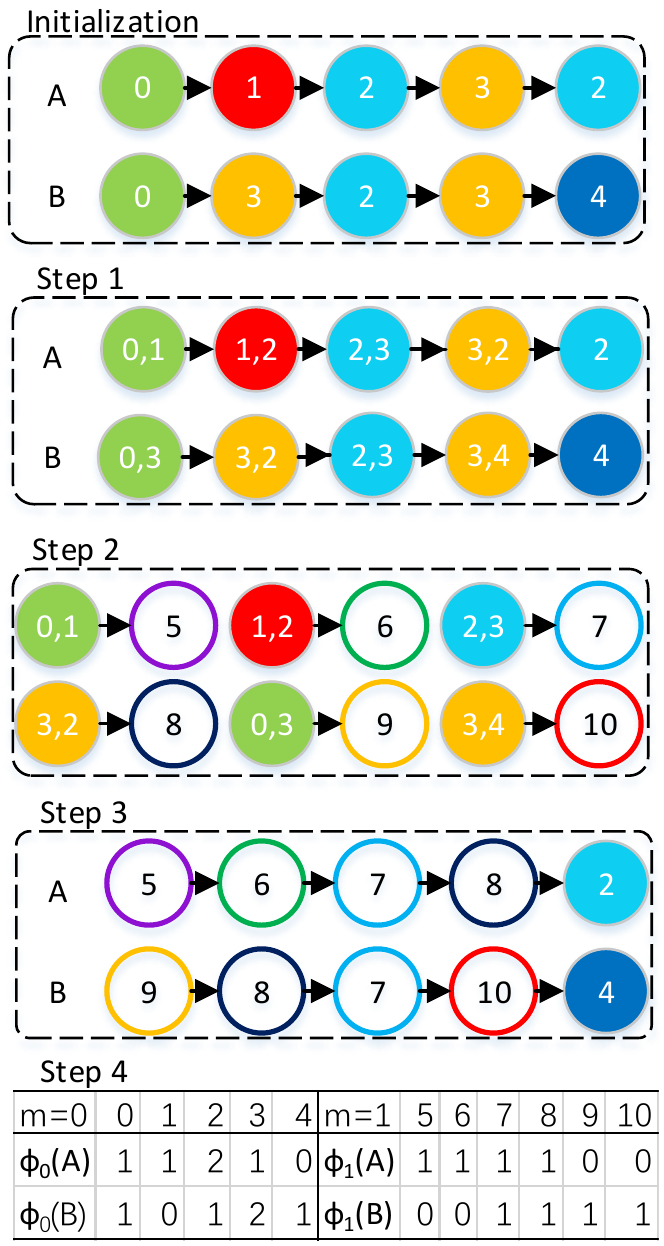}
    \caption{WL kernel illustration. 
    At initialization, there are two pruning proposals with features at $m=0$.
    At step 1, WL kernel collects the next labels of each node.
    At step 2, it re-encodes or re-labels the new nodes incorporating their neighbour information.  
    At step 3, it obtains a new graph with features at $m=1$.
    At step 4, WL kernel compares  the histogram on both $m=0$  and  $m=1$ features.
    The iteration repeats until $m=M$.}
    \label{fig: kl}
\end{figure}

We illustrate the WL kernel in Fig. \ref{fig: kl}. As shown in the figure,  at initialization, there are two pruning proposals A and B with  features of $m=0$.
At step 1, WL kernel collects the next label of each node. For example, for proposal A, the next node of 0 is 1, so we put the information of the next node into the original node and obtain (0,1).
At step 2, it re-encodes or re-labels the new nodes incorporating their neighbour information. For example, (0,1) is relabeled as 5.   
At step 3, it obtains a new graph with features at $m=1$.
At step 4, WL kernel compares  the histogram on both $m=0$  and  $m=1$ features. The histogram is the number of each node in the graph. 
The iteration repeats until $m=M$.

\section{Acceleration Techniques} \label{app: accelerate}
We adopt parallel updates,  early stopping and parallel compiler optimization to accelerate the scheme evaluation.

\paragraph{Parallel updates.} 
As pruning a DNN model can take hours, a distributed pruning and asynchronous parameter updating method is adopted  in order to speed up the learning of the controller \cite{dean2012large}. 
More specifically, a parameter-server scheme is created to store the shared parameters of the generator RNN replicas. Each generator replica samples $B$ different unified schemes which are evaluated  in parallel. When the evaluation finishes, their performance are recorded and the generator can compute gradients. The gradients are then sent back to the  parameter server such that all the generator replicas can be updated. 

\paragraph{Early stopping.} 
For each pruning scheme, the pruning usually have two phases: pruning phase and fine-tuning phase. The pruning phase focuses on finding out what weights should be pruned in each layer and the fine-tuning phase tries to improve the mAP problem while keeping the pruned weighs zero. In previous works, it normally costs a lot of training epochs for both phases to get the best pruning performance. However, in our work, we use early stopping strategy to reduce the number of training epochs in each pruning implementation. The training set is split and 10\% training data are formulated as a validation set which are never used for training.   The loss on the validation set after each  training epoch is monitored and we stop training when the monitored validation loss does not get improved for 3 epochs.

\begin{table*}[tb]
\center
\caption{Comparison of 3D detection methods}
\begin{center}
\begin{threeparttable}
\scalebox{1.1}{
\begin{tabular}{c|c|c|c|c|c|c|c|c|c}
\toprule
\multirow{2}{*}{ Detection methods} & \multirow{2}{*}{Modality} & \multirow{2}{*}{ \makecell{ Server GPU\\ speed (ms)}} & \multirow{2}{*}{ \makecell{ Mobile GPU\\ speed (ms)}}  & \multicolumn{3}{c|}{Pedestrian 3D detection} & \multicolumn{3}{c}{Cyclist 3D detection} \\
\cline{5-10}
                         &                           &       &                 & Easy      & Moderate      & Hard     & Easy      & Moderate      & Hard      \\
\midrule

           F-PointNet \cite{qi2018frustum}       &      \multirow{3}{*}{R+L\tnote{a} }                   &    170    &   -       &    51.21  & \textbf{44.89 } & 40.23 & 71.96 &  56.77 &  50.39 \\
           AVOD-FPN \cite{ku2018joint}  &           & 100      &   - &      50.80  & 42.81 &  40.88 &  64.00 &  52.18  & 46.61     \\
           UberATG-MMF \cite{liang2019multi}   &       &      80    &-  &      -& -& - & - & -& -        \\
           \midrule
           PointPillars   \cite{lang2019pointpillars}    &      \multirow{5}{*}{L }      &       20   &  260 &       52.08 & 43.53 & 41.49 & 75.78 & 59.07 & 52.92  \\
           SECOND \cite{yan2018second}  &    &   50 & - &  51.07 & 42.56 & 37.29 & 70.51 & 53.85 & 46.90 \\
           Point-GNN \cite{shi2020point} &    &    643  & -  &  51.92 & 43.77 & 40.14 & \textbf{78.60} & \textbf{63.48} & \textbf{57.08} \\
           Ours     &     &  \textbf{ 18 }  &   97   &    \textbf{52.33} & 43.68 &  \textbf{41.52 }&  76.09  & 59.31 & 53.56
 \\
\bottomrule
\end{tabular}}
 \begin{tablenotes}
 \item[a] \small  'L' and 'R' represent  LiDAR and RGB images respectively; \\
 \end{tablenotes}
\end{threeparttable}
\end{center}
\label{tab: ped}
\end{table*}

\paragraph{Parallel compiler optimization.} 
The code generation framework with compiler optimization is adopted to actually measure the inference speed on mobile devices for more accurate speed measurement. The code generation framework supports all kinds of pruning types with the ability to dynamically auto-switch between these pruning types. The speed measurement with compiler optimization usually takes hours. However, it can be performed in parallel with the mAP evaluation step in the actor and does not incur extra time cost. Note that to measure the speed, we do not need the best pruned and  well-trained model weights. As long as the pruned localizations are determined, we can start the speed measurement with compiler optimization while in the meantime the pruned model can be retrained to improve its mAP  during evaluation.

\section{Performance for Pedestrians and Cyclists} \label{app: pedestrian}

We show the performance of pedestrians and cyclists in Tab. \ref{tab: ped}. We can observe that the proposed method can achieve real-time inference on mobile devices with state-of-the-art detection performance. As the server GPU (GTX 1080Ti) is more powerful than the mobile GPU (GPU on Samsung Galaxy S20 phone), the inference speed on server GPU is faster than that on mobile GPU. Besides, since other 3D detection methods use various specific layers and structures, such as sparse 3D CONV layers, leading to highly irregular memory and computation pattern, currently these methods are not supported by the compiler optimization. Moreover, it is also challenging to support these methods with compiler optimization as the irregular patterns are hard to optimize.  
Although there are some methods with high mAP such as Point-GNN \cite{shi2020point}, they usually  take much longer time (e.g. 643ms for Point-GNN) to process one LiDAR image on average on server GPUs.
It is even harder to implement them on mobile GPUs considering the computation resource gap.  
Although  the  PointPillars \cite{lang2019pointpillars}  is the fastest on server GPUs, it still  takes about 260ms on mobile GPU, which can hardly satisfy the real-time requirement. However, with our proposed method, we can use only 97ms to process one LiDAR image on mobile GPUs, achieving real-time inference on mobile.

\section{Experimental Setting Details} \label{app: exper}

All experiments are conducted with the KITTI object detection benchmark dataset \cite{Geiger2012CVPR}, which consists of samples with both lidar point clouds and images. We follow the standard convention  \cite{zhou2018voxelnet} of only using lidar points. 
The dataset are originally divided into 7481 training and 7518 testing samples. For experimental studies
we split the official training into 3712 training samples and
3769 validation samples.
We train a PointPillars model  for cars  use a pillar resolution of 0.24m with   12000 as the max number of pillars  and 100 as
the  max number of points per pillar.
During inference, we apply axis aligned non maximum suppression (NMS) with an overlap threshold of 0.7 intersection-over-union (IoU) following the official KITTI  protocol.

The generator RNN is a two-layer LSTM with 49 hidden units on each layer. Its weights  are initialized uniformly between -0.1 and 0.1. The ADAM optimizer \cite{kingma2019method} is employed with a learning rate of 0.0005. 
For the parallel training,  we  use 50 GPUs to concurrently evaluate the unified schemes. The generator training takes about 10 days. 

During the evaluation, for  each unified scheme, we first perform intra-kernel replacement and fine-tuning the model with 5 epochs. Then we start to prune the model. For ADMM pruning, we prune with 5 epochs and fine-tune the pruned model with 10 epochs. For magnitude pruning, we first prune the model according to weight magnitudes and then fine-tune the model with 10 epochs. 
Note that we start from a well-trained model and thus it does not need too many epochs to obtain competitive results with network enhancement and pruning. We use the ADAM Optimizer \cite{kingma2019method}
with a learning rate of 0.0002, weight decay of 1e-4 and momentum of 0.8. 

For the reward function (Eq. \eqref{eq: reward}), we set $\alpha$ to 0.01 and the inference time and threshold are measured in millisecond to incur large penalty so that the speed requirement is more significant than the mAP improvement. 

For the Bayesian optimization, we set the  Bayesian batch size $B = 10$ (i.e., we select unified schemes with top 10  acquisition function values and  evaluate them in parallel) with a pool size $K=50$.

\section{Ablation Study} \label{app: ablation}

\begin{table}[tb]
\caption{Comparison of different grid sizes}
\begin{center}
\begin{threeparttable}
\scalebox{0.84}{
\begin{tabular}{c|c|c|c|c|c}
\toprule
\multirow{2}{*}{ \makecell{Pruning \\ methods}}  & \multirow{2}{*}{\makecell{ server GPU\\ Speed  (ms)}}  & \multirow{2}{*}{\makecell{ mobile GPU \\ Speed  (ms)}}  & \multicolumn{3}{c}{Car 3D detection} \\
\cline{4-6}
            &    &  & Easy      & Moderate      & Hard        \\
\midrule
        PointPillars \cite{lang2019pointpillars} (0.16)   & 25  & 542 & 84.67 & 75.11 & 69.53 \\
        PointPillars \cite{lang2019pointpillars} (0.24)    & 20 &  257 & 84.05 & 74.99 & 68.30 \\
        PointPillars \cite{lang2019pointpillars} (0.32)   & 18 &  223 &  80.81 & 69.48 & 67.19 \\
Ours (0.16)   &  24 & 189 & 85.50 & 76.58 & 70.58  \\
Ours (0.24)   & 18 & 97 & 85.20  &  75.57 & 68.37  \\
   Ours (0.32)   & 17 &  85 & 80.98 & 70.14 & 67.37 \\
\bottomrule
\end{tabular}}
\end{threeparttable}
\end{center}
\label{tab: grid}
\end{table}

We explore the effects of various pillar grid sizes and show the performance in Table \ref{tab: grid}. We can observe that larger grid size results in smaller pseudo-image input image sizes, thus the inference speed can be improved at the 
cost of detection performance degradation. For the grid size of 0.32m, the performance degradation is non-neglectable and thus we do not use the grid size of 0.32m.

We test the case with pruning search only instead of unifying network enhancement and pruning search as shown in Tab. \ref{tab: ablation}. Under the same configuration, pruning search only can achieve a mAP of 74.89\% with an average inference time of 108ms. Compared with the unified framework with 76.38\% mAP and 97ms inference, we can see that incorporating network enhancement can further improve the detection and speed performance.  We also note that although the real-time threshold is set to 100ms, without Winograd, pruning search still slightly violate the real-time requirement, demonstrating the acceleration performance of Winograd. 

Moreover, as we increase the real-time requirement from 100ms to 90ms, the proposed method achieves a mAP of 74.15\% with an average 89ms speed, demonstrating that the method prunes the model harder to improve speed, thus incurring detection performance degradation. We show the results in Table \ref{tab: ablation}.

\begin{table}[tb]
\caption{Comparison of various methods}
\begin{center}
\begin{threeparttable}
\scalebox{0.94}{
\begin{tabular}{c|c|c|c|c}
\toprule
\multirow{2}{*}{ \makecell{ methods \\ (real-time threshold)}}   & \multirow{2}{*}{\makecell{ mobile GPU \\ Speed  (ms)}}  & \multicolumn{3}{c}{Car 3D detection} \\
\cline{3-5}
               &  & Easy      & Moderate      & Hard        \\
\midrule
pruning search only (100ms)    &  108 & 83.92 & 74.12 & 66.63 \\
Ours (100ms)  & 97 & 85.20  &  75.57 & 68.37  \\
Ours (90ms) & 89 & 82.33  &  73.60 & 66.52  \\
\bottomrule
\end{tabular}}
\end{threeparttable}
\end{center}
\label{tab: ablation}
\end{table}

\end{document}